\documentclass{article}

\usepackage[preprint]{neurips_2024}
\usepackage{amsmath}

\usepackage[table]{xcolor}         
\definecolor{royalblue}{RGB}{65, 105, 225}

\usepackage[utf8x]{inputenc} 
\usepackage[T1]{fontenc}    
\usepackage[hidelinks, 
            colorlinks = true,
            allcolors = royalblue
        ]{hyperref}  
\usepackage{url}            
\usepackage{booktabs}       
\usepackage{amsfonts}       
\usepackage{nicefrac}       
\usepackage{microtype}      
\usepackage{graphicx}
\usepackage{subcaption}
\usepackage{xcolor}
\usepackage{listings}
\usepackage{tcolorbox} 
\usepackage{subcaption}

\title{\textsc{WeatherFormer}: A Pretrained Encoder Model for Learning Robust Weather Representations from Small Datasets}
\usepackage{longtable}

\newcommand{\wf}{\textsc{WeatherFormer}}

\author{%
  Adib Hasan \\
  MIT\\
  \texttt{notadib@mit.edu} \\
  \And
  Mardavij Roozbehani\\
  MIT\\
  \texttt{mardavij@mit.edu}
  \And
  Munther Dahleh \\
  MIT\\
  \texttt{dahleh@mit.edu}
}

\begin{document}

\maketitle

\begin{abstract}
   This paper introduces \wf{}, a transformer encoder-based model designed to learn robust weather features from minimal observations. It addresses the challenge of modeling complex weather dynamics from small datasets, a bottleneck for many prediction tasks in agriculture, epidemiology, and climate science. \wf{} was pretrained on a large pretraining dataset comprised of 39 years of satellite measurements across the Americas. With a novel pretraining task and fine-tuning, \wf{} achieves state-of-the-art performance in county-level soybean yield prediction and influenza forecasting. Technical innovations include a unique spatiotemporal encoding that captures geographical, annual, and seasonal variations, adapting the transformer architecture to continuous weather data, and a pretraining strategy to learn representations that are robust to missing weather features. This paper for the first time demonstrates the effectiveness of pretraining large transformer encoder models for weather-dependent applications across multiple domains.
\end{abstract}

\section{Introduction}
Understanding dynamic weather patterns is extremely important in several major fields including agriculture, epidemiology, climate science, disaster response, and transportation. However, real-world datasets in these fields are small and lack detailed weather measurements. For instance, a crop yield prediction dataset typically contains only 5-7 years of detailed weather data for a few farms \citep{McFarland2020g2f} or just a few weather measurements over a long period \citep{khaki2019cnn}. Consequently, large models with sufficient capacity to learn weather patterns overfit when trained on these datasets. On the other hand, without a good representation of weather, any model's performance for prediction tasks in these fields will remain sub-optimal.

In Natural Language Processing (NLP), the problem of small datasets is tackled by training a large model on a massive unlabelled text dataset, such as the English Wikipedia \citep{devlin2019bert}, and then finetuned on small datasets for prediction tasks. This approach with pretrained large models like BERT \citep{devlin2019bert}, RoBERTa \citep{liu2019roberta}, and others \citep{clark2020electra, lan2020albert} has demonstrated remarkable success in improving the benchmarks on sentiment analysis \citep{Batra_2021}, machine translation \citep{zhu2020incorporating}, and reading comprehension \citep{fernandez2023automated}. 

Unlike text, which is a discrete domain, weather data is continuous and has both spatial and temporal dependencies. Hence, it is unclear if large models pretrained on public weather datasets will improve the performance of the downstream tasks. While recent works such as \citet{man2023wmae} and \citet{nguyen2023climax} have proposed decoder-based \citep{vaswani2017} transformer models for weather forecasting and downsampling, no foundational weather model to our knowledge has been trained to extract good representations of weather from a small number of observations. 

In this paper, we aim to fill this knowledge gap by pretraining a weather \textit{encoder model}, called \wf{} on a large dataset of satellite-based weather measurements from the NASA Power Project \citep{NASAPower}. Our model is trained to extract good representations of weather from a small dataset and just a few basic measurements such as the mean temperature and precipitation. It supports a maximum sequence length of 365, allowing it to process up to 1 year of daily weather data, 7 years of weekly weather data, and 30 years of monthly weather data. It also incorporates a new positional encoding mechanism sensitive to geographical location, year, and seasonality, thus enabling it to capture weather patterns' dynamic and cyclic nature across different times and places. We also observed that Masked Language Modeling (MLM) \citep{devlin2019bert}, a common pretraining strategy for encoder models, is suboptimal for pretraining large models with continuous weather data. For a remedy, we propose a novel pretraining strategy to address this shortcoming. 

\begin{figure}[htb]
    \centering
    \includegraphics[width=0.55\textwidth]{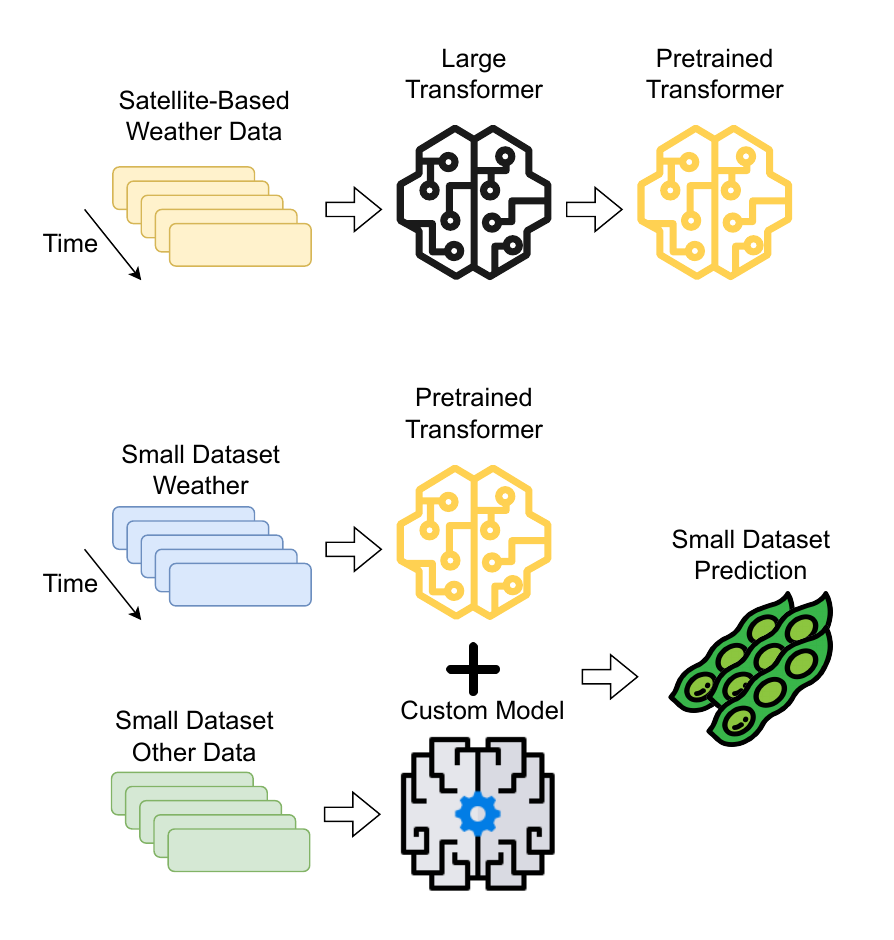}
    \caption{A large transformer model (\wf{}) is pretrained on a massive satellite-based weather dataset, enabling the model to learn rich representations of weather during pretraining. This pretrained model can extract robust weather features for a new prediction task (Small Dataset). Even when only a limited number of weather measurements are available for a specific task, the learned weather features can be used to improve the prediction accuracy.}
    \label{fig:big-picture}
\end{figure}

The potential impact of a pretrained encoder weather model such as \wf{} is substantial across various domains, including agriculture, epidemiology, climate science, and transportation. Our experiments demonstrate that fine-tuning this model for soybean yield prediction and influenza forecasting in New York City achieves state-of-the-art performance in both tasks. However, the applications extend far beyond these examples. In agriculture, additional applications could be predictions for crop yield, plant diseases, and flowering periods, which are crucial for resource-efficient farming. In epidemiology, the outbreaks of weather-dependent diseases besides influenza, such as dengue \citep{Abdullah2022}, malaria \citep{Hoshen2004}, cholera \citep{Christaki2020Cholera}, and typhoid \citep{Jia2024Typhoid}, can be predicted. In climate science, \wf{} can be utilized to study various environmental phenomena such as droughts \citep{Balting2021}, coastal flooding \citep{Kirezci2020}, and soil erosion \citep{EEKHOUT2022soilerosion}.





Our contributions can be summarized as follows: 

\begin{itemize}
    \item Collected, processed and created a 60 GB dataset of satellite weather measurements ready for training large deep learning models.
    \item Modified the transformer encoder model with a novel positional encoding, scalers, and a novel pretraining task to pretrain on a large volume of continuous weather data.
    \item Trained and open-sourced two foundational models for weather, called \wf{}, with 2 million and 8 million parameters, respectively.
    \item Finetuned \wf{} models to achieve SOTA performance in Soybeans yield prediction in the US corn belt and Influenza forecasting in New York City.
\end{itemize}
\section{Background and Related Work}
\subsection{Foundational Models in Deep Learning}
Foundational models, exemplified by GPT \citep{radford2018improving} and BERT \citep{devlin2019bert}, mark a transformative shift in the development and deployment of deep learning systems in Natural Language Processing (NLP). These models are pre-trained on extensive data corpora and demonstrate an exceptional capacity to generalize across diverse tasks without task-specific training. 

The foundational models have also found applications beyond NLP. In meteorology, foundational models have been introduced for short and medium-term weather prediction, downsampling, and climate-related text analysis. \citet{pathak2022fourcastnet} developed a time series-based recurrent neural network \citep{hochreiter1997long} for short to medium-range weather forecasting. \citet{lam2023graphcast} employed Graph Neural Networks \citep{scarselligraph2008} to process satellite imaging data for global medium-range weather predictions. More recently, transformer-based models have been deployed for weather forecasting \citep{man2023wmae, nguyen2023climax} and data downsampling \citep{nguyen2023climax}. Lastly, researchers have developed both encoder-based \citep{webersinke2022climatebert, fard2022climedbert} and decoder-based models \citep{bi2024oceangpt} for analyzing climate-related textual data. To the best of our knowledge, no foundational weather model besides \wf{} has been trained to extract weather representations for downstream prediction tasks.

\subsection{Self-Supervised Learning}
Self-Supervised Learning (SSL) trains foundational models on unlabeled datasets using specialized tasks that do not require explicit labels, allowing models to autonomously extract useful information from the data.

\textbf{Masked Language Modeling (MLM):} Introduced by BERT, MLM involves masking certain tokens in a sentence and tasking the model with predicting these tokens using the visible context. The loss function for MLM is formulated as:
\[
\mathcal{L} = -\log P(\mathcal{M} | \mathcal{V}, \Theta)
\]
where $\mathcal{M}$ denotes the masked tokens and $\mathcal{V}$ the visible tokens.

\textbf{Autoregressive Language Modeling:} Popularized by GPT \cite{brown2020language}, this approach predicts the next token in a sequence based on the previous tokens, training the model to understand the probability distribution of a language. It is defined mathematically as:
\[
\mathcal{L} = -\sum_{k>1}^n \log P(x_k | x_1, \dots, x_{k-1}, \Theta)
\]
Where $x_1,\dots,x_{k-1}$ are previously generated tokens. This technique has also proven effective for modeling continuous weather data \citep{nguyen2023climax}.

Other notable SSL tasks include Replaced Token Detection \citep{clark2020electra} and Next Sentence Prediction \citep{devlin2019bert}.

\subsection{Machine Learning-based Yield Prediction}
Researchers have applied various machine learning methods to predict crop yields from wheat to potatoes using data ranging from on-field measurements to satellite imagery. Farm-level data contains detailed features such as weather conditions, soil characteristics, fertilization, and irrigation, enabling accurate yield predictions \citep{russ2010regression, ahamed2015applying}. In contrast, satellite data lacks such granularity and is generally used for broader yield predictions at regional, state, and national levels \citep{khaki2019cnn}.

Key algorithms in this domain include neural networks \citep{russ2008data, khaki2019cnn}, regression models \citep{russ2010regression}, random forests \citep{jeong2016random}, and clustering algorithms \citep{ahamed2015applying}. 

\subsection{Influenza Forecasting}
Influenza viruses cause significant annual epidemics worldwide, with the 2023-2024 season alone resulting in over 380,000 hospitalizations and 24,000 deaths \citep{cdcWeeklyUS}. To combat this, researchers have developed forecasting models that can be categorized into two main types: parametric machine learning-based models and neural network-based models.

Parametric models employ statistical learning techniques to fit epidemiological data for influenza epidemics. Examples include Gaussian Process regression \citep{zimmer2020flu}, Dynamic Bayesian models \citep{osthus2019dynamic}, and the Dante system \citep{osthus2021multiscale}. Random forest algorithms have also been applied for influenza activity prediction in Eastern China's subtropical zones \citep{liu2019influenza}.

Neural network-based models, particularly those leveraging recurrent neural networks, are used for capturing temporal and spatial dependencies in epidemic data. These models have been successful in predicting weekly influenza trends globally and in the U.S. \citep{wu2021deepgleam, amendolara2023lstm}. Additionally, recurrent networks have been utilized in China using climate, demography, and search engine data for forecasting influenza-like illnesses \citep{yang2023deep}.

\section{Data Collection}
\subsection{Pretraining Dataset}

Our pretraining data was downloaded from the NASA Power API \citep{NASAPower}. We downloaded 28 daily weather measurements from 1984 till 2022. Although the Power API has kept records of weather since 1980, we have found that the earlier years were missing many of the important weather measurements and we chose to exclude them. The dataset consisted of rectangular regions of shape $5^\circ\times8^\circ$ spanning the continental United States, Central America, and South America as shown in \autoref{fig:pretraining-loc-map}. The dataset contained 119 rectangles, each containing 160 unique coordinates with weather measurements. The spatial resolution of the dataset was $0.5$ degree$^2$. Additionally, we computed weekly and monthly averages and concatenated with the dataset. In total, the pretraining dataset contains approximately 9.1 billion floating point numbers. 

A detailed description of the dataset is given in \autoref{sec:datasets}. 
\begin{figure}[htb]
    \centering
    \begin{subfigure}{0.44\textwidth}
        \includegraphics[width=\textwidth]{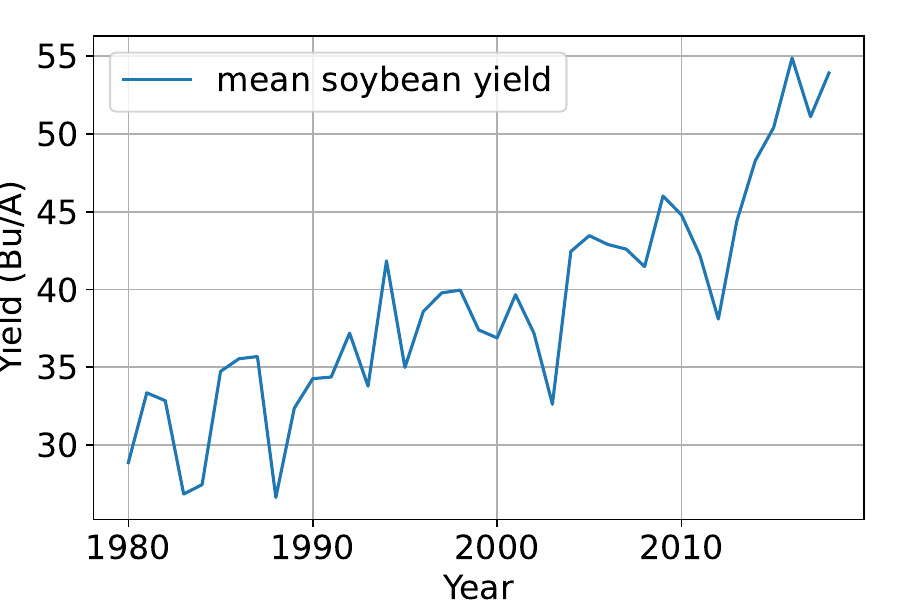}
    \caption{Mean soybean yield (Bu/A) across 9 corn belt counties of the United States.}
    \label{fig:mean-soybean-yield}
    \end{subfigure}
    \hfill
    \begin{subfigure}{0.53\textwidth}
        \includegraphics[width=\textwidth]{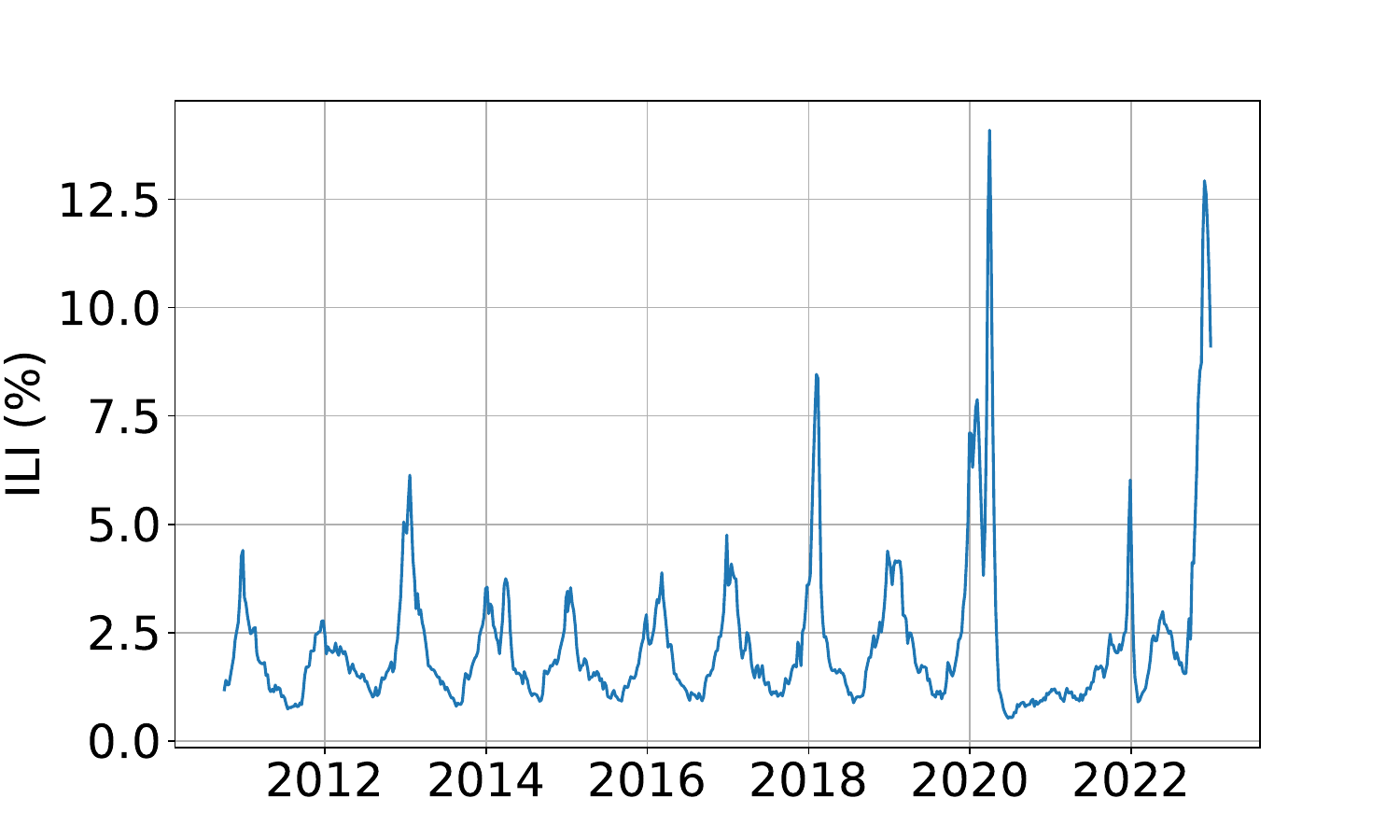}
    \caption{Influenza Like Illness ($\%$) for New York City from 2010 to 2022. }
    \label{fig:nyc-flu-cases}
    \end{subfigure}
    \label{fig:finetuning-datasets}
    \caption{County-level soybean yield and Influenza forecasting for New York City were used as the finetuning tasks for \wf{}. The mean soybean yield gradually increased due to hybrid vigor and better farming practices. On the other hand, the influenza seasons show clear peaks during the winter until 2019 and after that, the patterns became irregular due to COVID-19.}
\end{figure}
\subsection{Finetuing Datasets}
To evaluate the downstream task performance of \textsc{WeatherFormer}, we predicted soybean yield in the corn belt region of the United States from 1980 to 2022 and also Influenza cases in New York City from 2010 to 2020. Following the conventions from the previous research, we reported Root Mean Square Error in Bu/Acre for the soybeans dataset and Mean Absolute Error (MAE) for the influenza dataset. A detailed description of the fine-tuning datasets, preprocessing, and prediction tasks can be found in \autoref{sec:datasets}.

\section{Architecture}
\begin{figure}[htb]
    \centering
\includegraphics[width=0.9\textwidth]{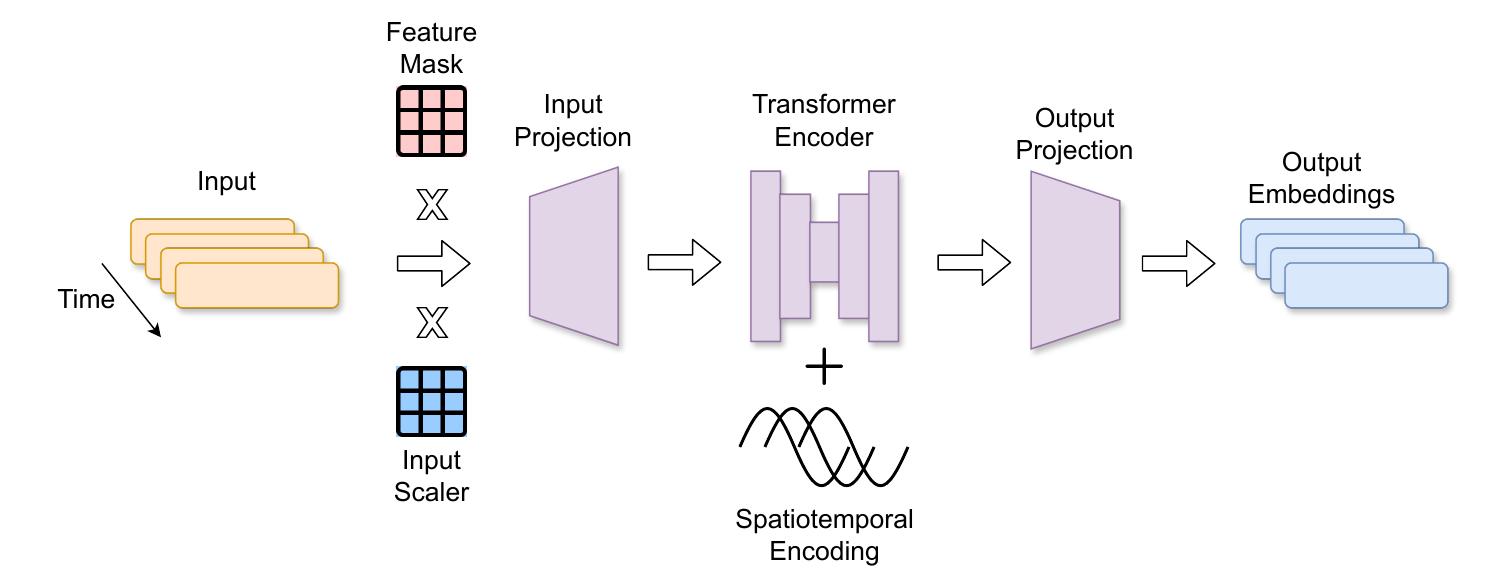}
    \caption{The forward pass of the weather inputs through the \wf{} architecture. The input is first multiplied with learnable input scalers and a feature mask and then projected to a hidden dimension through a linear layer. After that, the input goes through a transformer encoder with a novel spatiotemporal encoding mechanism and finally, the input is projected to an output dimension.}
    \label{fig:weatherformer-arch}
\end{figure}
In this section, we describe the architecture of the \wf{} in detail. \autoref{fig:weatherformer-arch} shows a visualization of the forward pass through the architecture. 
\subsection{Feature Mask and Padding Mask}
The \textsc{WeatherFormer} model expects an input of the shape $N\times 31$, where $N$ is the length of the input sequence and $31$ is the number of weather measurements. \wf{} allows some features to be missing during finetuning. For this reason, there is a feature mask that fills the missing features with zeros. There is also a padding mask to allow \textsc{WeatherFormer} to process variable length inputs, however, the maximum context length during pretraining was 365.

\subsection{Scaling Parameters}
We introduce learnable scaling parameters for each weather input dimension since standardized weather measurements (mean 0, std. dev. 1) might not align with the fixed range of the sinusoidal positional encodings ($[-1, 1]$). Having scaling parameters ensures that the positional encodings contribute meaningfully to the learned representations of weather data, allows the model the differentially rank the importance of the weather features, and integrates temporal scales (daily, weekly, monthly) directly into the model architecture through distinct scaling embeddings.

We implement scaling with a PyTorch Embedding layer, assigning different embeddings for temporal granularity of input between 1 and 30. All embeddings are initialized with 1 and the embeddings for 1, 7, and 30 (corresponding to daily, weekly, and monthly input data) are learned during pretraining.

\subsection{Spatiotemporal Positional Encoding}
The input of \wf{} is a time series of weather measurements, which has both temporal and spatial dependency. The original sinusoidal positional encoding of the transformers will be unable to take this into account. For this reason, we designed a new Spatiotemporal Positional Encoding for the \textsc{WeatherFomer}:
\begin{align*}
    PE(\mathrm{year}, \mathrm{lat}, \mathrm{lng})_{\mathrm{pos}, 4i} &= \sin\left(\mathrm{pos}\cdot\cdot 10000^{-4i/d}\right)\\
    PE(\mathrm{year}, \mathrm{lat}, \mathrm{lng})_{\mathrm{pos}, 4i+1} &= \cos\left(\mathrm{pos}\cdot\cdot 10000^{-4i/d}\right)\\
    PE(\mathrm{year}, \mathrm{lat}, \mathrm{lng})_{\mathrm{pos}, 4i+2} &= \sin\left(\frac{\pi\cdot\mathrm{lat}}{180}\cdot\cdot 10000^{-4i/d}\right)\\    
    PE(\mathrm{year}, \mathrm{lat}, \mathrm{lng})_{\mathrm{pos}, 4i+3} &= \cos\left(\frac{\pi\cdot\mathrm{lng}}{180}\cdot\cdot 10000^{-4i/d}\right)\\    
\end{align*}
The scaler $\pi/180$ converts the coordinates into radians and also ensures in every $360^\circ$ distance, the spatial encoding repeats itself. 

\subsection{Transformer Encoder and Output Projection}
Once both the temporal granularity encodings and the spatiotemporal encodings are added to the input, the input is then passed through a transformer encoder. After the transformer encoder, the model is finally projected to an output dimension. For an input of the shape $N\times 31$, the model produces an output of the shape $N\times M$, where $M$ is the desired output dimension.

\section{Pretraining}
We pretrained two models of size 2M and 8M, respectively. Each model was pre-trained with the task of predicting 10 weather variables based on the remaining 21. To do so, 10 target weather variables were masked with the input mask. Then the model learned to predict the missing values. We chose mean square error as our loss function. In every batch, one input and one output variable were swapped, and consequently, over time the model learned to predict every weather variable from the combination of other variables. 

By training the model to predict missing weather variables on a pretraining dataset, we encourage the development of robust representations derived from partial observations, making it well-suited for our downstream tasks.

\textbf{Optimization:} We pretrained each model for 75 epochs over the training data and computed performance over the validation dataset after every epoch. Adam \citep{kingma2017adam} was used as the optimizer with a learning rate of $0.0005$, a warm-up period of 10 epochs, a batch size of 64, and an exponential learning rate decay factor of $0.99$. These hyperparameters were selected after pretraining small-scale models during other preliminary investigations. We did not explore the full search space for the full-scale models due to the prohibitive expansiveness of the task. The 2M and the 8M parameter models took 17 hours and 39 hours to train on two NVIDIA V100 GPUs.

\section{Finetuning}
We fine-tuned the \wf{} model to predict county-level soybean yield and influenza-like illness (ILI) percent for New York City. Both of these tasks are influenced by weather, but other features are also important. For instance, crop yield is influenced by soil profile, management practices, and past yield. On the other hand, ILI on \autoref{fig:nyc-flu-cases} show a clear autoregressiveness. Therefore, for prediction tasks, we processed the weather with \wf{} and the rest of the features with another model and then combined them together to predict the target variable. A sample model for this approach is shown in \autoref{fig:big-picture}.

\subsection{Soybean Yield Prediction}
The soybeans yield prediction dataset has weather measurements, soil characteristics, management practices, and past yield data. We fine-tuned \wf{} model in this dataset in two ways as described below:

\begin{figure}[htb]
    \centering
    \includegraphics[width=0.98\textwidth]{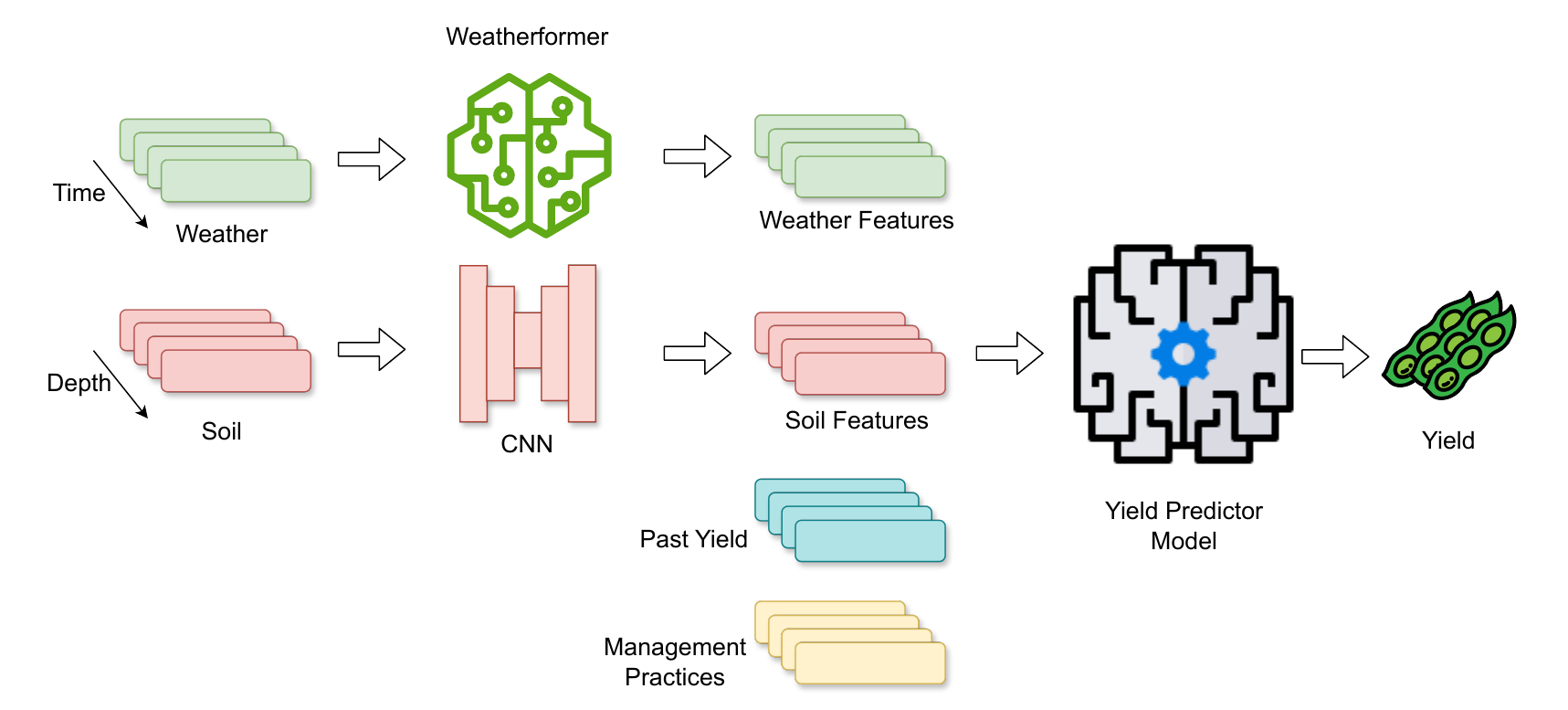}
    \caption{Soybean yield predictor architectures utilizing \wf{}. The weather measurements for the current year and the last few years are processed with \wf{} and the soil measurements are processed with a CNN reported in \citet{khaki2019cnn}. Then the yield is predicted with either a linear layer or a transformer. The entire model is trained at once. Since yield for the current year is the target variable, it is replaced in the input with last year's yield.  }
    \label{fig:yield-predictors}
\end{figure}

\textbf{\wf{}+Linear Model:} Weekly weather features for the current year and the past two years are processed with \wf{} separately and the dimension is reduced to 120 for each year. Next, the soil properties are processed with the CNN used in \citet{khaki2019cnn}. These features are concatenated with management practices and past yields for each year. Since the current year's yield is unknown, it is replaced by the last year's yield. A single linear layer predicts yield from this input. This model highlights \wf{}'s direct impact but is suboptimal for yield prediction since the autoregressive pattern of soybean yield as shown in \autoref{fig:mean-soybean-yield} is not taken into account.

\textbf{\wf{} + Transformer:} We replaced the last linear layer of the previous model with a transformer to capture the autoregressive trend of the data and then predicted the yield. We observed that 7 years of data gives the optimal performance on the validation set as opposed to 3 in the previous model.

\textbf{Baseline Models:} We tested three baseline models: a least squares linear regression model, the CNN-RNN model as proposed by \citet{khaki2019cnn}, and a novel CNN-Transformer model. 

The CNN-RNN model uses two separate CNNs to extract features from soil data across the depth dimension and weather data across the time dimension for the last three years. These features are then concatenated with the remaining data and passed through an LSTM to capture autoregressive patterns and then yield is predicted. 

Next, we replaced the LSTM of the CNN-RNN model with a transformer encoder, since it has the self-attention mechanism and our novel spatiotemporal encoding can encode coordinates. This results in further improvement over the CNN-RNN model. We call this model CNN-Transformer in \autoref{tab:pretrained-soybeans-yield}.

\textbf{Optimization: } All experiments used Adam optimizer, batch size 64, initial learning rate of 0.0005, 10-epoch warm-up, followed by exponential decay (factor 0.95). Models were trained for 40 epochs and evaluated on the validation set with the root mean square error (RMSE). The optimum number for past years' data is different for each model and was optimized from 1 to 7 years. Each model took between 1 to 3 hours to train and evaluate on one NVIDIA V100 GPU.

\subsection{Influenza Forecasting}
\begin{figure}[htb]
    \centering
    \includegraphics[width=0.94\textwidth]{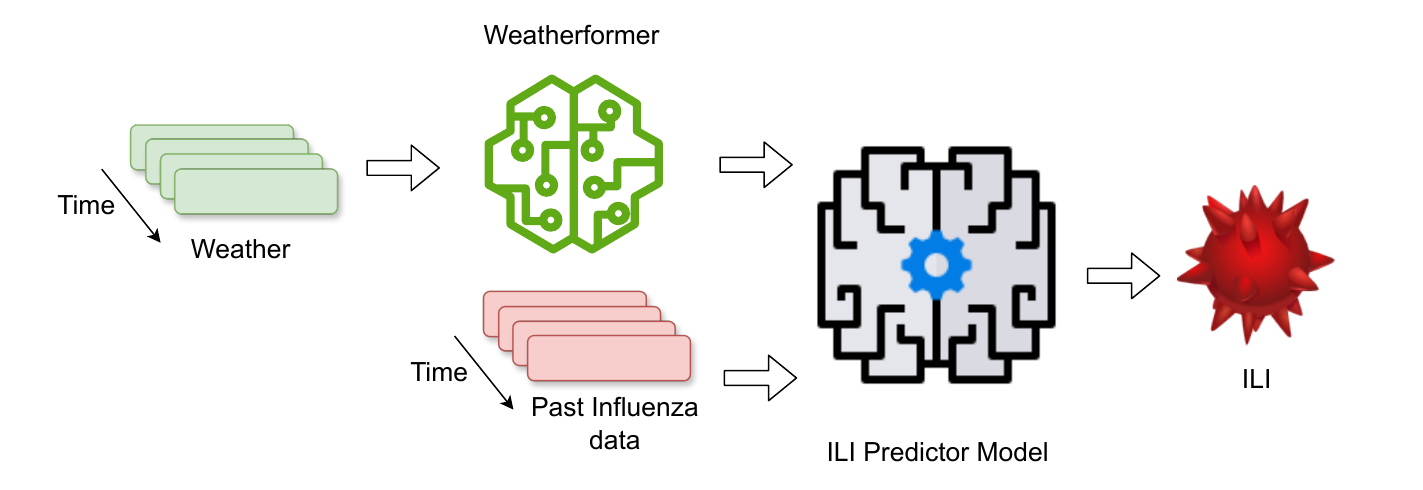}
    \caption{Influenza Like Illness (ILI) percent predictor architecture utilizing \wf{}. The weather measurements are first processed with \wf{} to extract useful features. These weather features and the past influenza data are processed by either a transformer to predict ILI percent for the next 10 weeks.}
    \label{fig:flu-predictor}
\end{figure}
\textbf{\wf{}+Transformer:} Like the yield prediction problem, we process the weekly weather data with the \wf{} and concatenate with the historical influenza data (influenza like illness and number of total patients). This is then passed through a second transformer as shown in \autoref{fig:flu-predictor} to predict the future influenza like illness (\%) for the next 10 weeks. The second transformer has three layers and a hidden dimension of 64. The model parameters were chosen after testing with hidden dimension sizes of 8, 16, 32, and 64, and 1-4 layers. We also observed that adding the last known ILI value to the first predicted value of the model improved its performance.

\textbf{Baseline Models:}: For our baselines, we trained a least square linear regression model, an Autoregressive Integrated Moving Average (ARIMA) \citep{Harvey1990ARIMA} model, and two transformer encoder models: one was trained only with the past ILI percentage and the total number of patients for timeseries analysis of the data. The other transformer model utilized both the past influenza data and the weather data. Training two transformer models allows us to quantify the effects of weather on ILI percentage forecasting. 

\textbf{Optimization:} Each model predicted the ILI percentage for a future period of 10 weeks. We used Mean Square Error (MSE) as the loss function for all the models, optimized using Adam optimizer with an initial learning rate of 0.0009 and for 30 epochs. We also employed an exponential learning rate scheduler that incorporated a 5-epoch warm-up phase followed by a decay factor of 0.95. All models took under 1 hour on one NVIDIA V100 GPU to train. Mean absolute error of 1, 5, and 10 weeks ahead predictions were reported in \autoref{tab:pretrained-flu-pred}.

\section{Results}
\begin{figure}[tbh]
    \centering
    \begin{subfigure}{0.45\textwidth}
    \includegraphics[width=\textwidth]{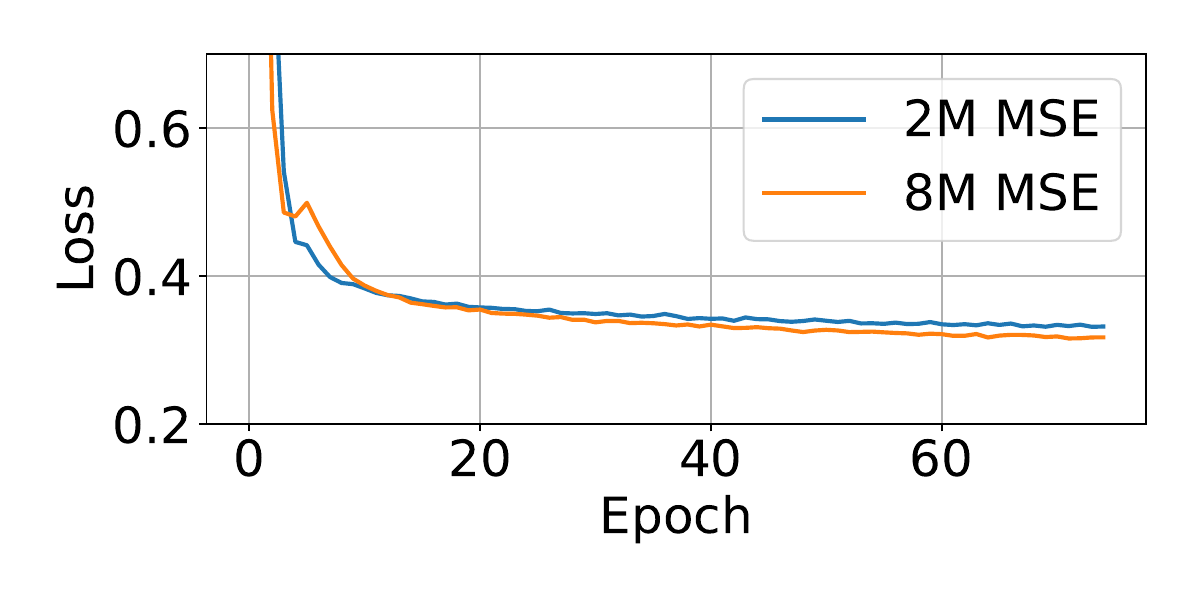}
    \caption{Training.}
    \label{fig:pretraining-train-loss}
    \end{subfigure}
    \begin{subfigure}{0.45\textwidth}
    \includegraphics[width=\textwidth]{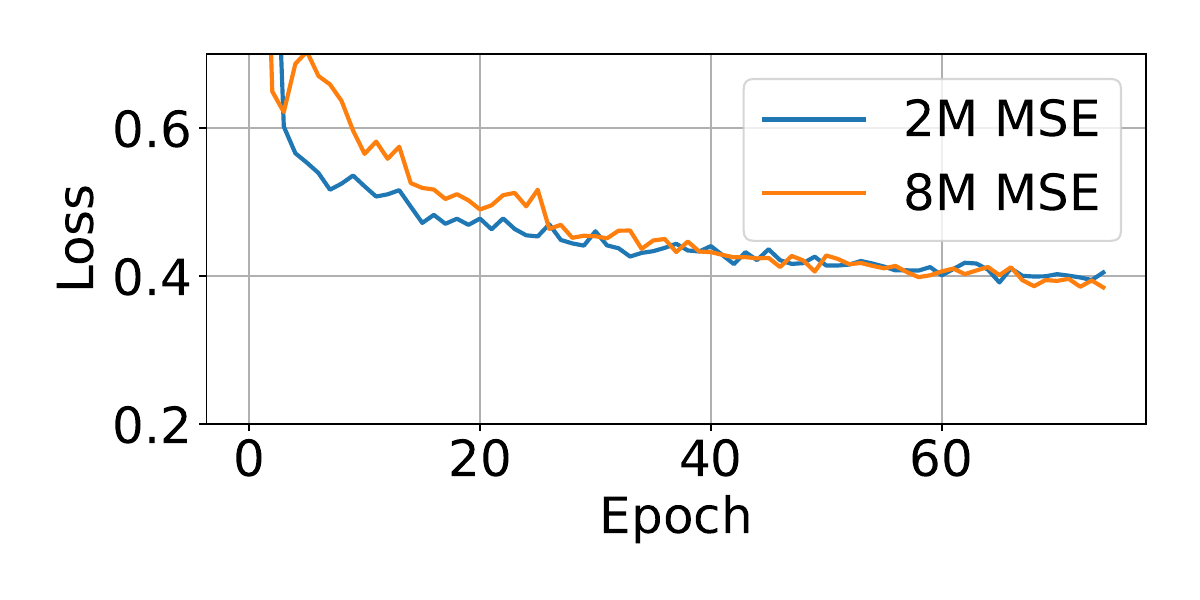}
    \caption{Validation}
    \label{fig:pretraining-val-loss}
    \end{subfigure}
     \caption{Comparison between the 2M and 8M model losses during pretraining. Both models performed similarly in the validation dataset.}
    \label{fig:pretraining-loss}
\end{figure}
\subsection{Pretraining}
We did not observe performance improvement with larger sizes, which could be because larger models require more data to be effective. We also observed that the input scaler weights shifted away from 1 during training. For the 2M parameter model, the mean value for the input scalers was 0.346, and the mean value for the weekly input scalers was 0.682, suggesting that the model learned to differentiate between data from different temporal granularities.

\subsection{Soybean Yield Prediction}
In \autoref{tab:pretrained-soybeans-yield}, we observe that the CNN-Transformer model performs better than the CNN-RNN model, suggesting the superiority of the transformer model to capture autoregressive patterns. On the other hand, \wf{}+Linear models show remarkable performance, due to superior weather feature extraction, but is still worse than the best model since autoregressive features were not captured well. Only when the \wf{} model and the transformer model are combined in \wf{}+Transformer models, the performance is optimum. Both the 8M and the 2M model performed similarly after finetuning.

\begin{table}[htb]
    \centering
    \caption{Comparison of Validation RMSE for county-level soybeans yield forecasting. The \wf{} models are pretrained and the mean and the standard deviation of the dataset are 38.5 Bu/Acre and 11.03 Bu/Acre, respectively.}
    \begin{tabular}{p{0.35\textwidth}l}
    \toprule
    \textbf{Model} & \textbf{Validation RMSE} \\\midrule
    Linear Regression & 6.90 \\
    CNN-RNN & 5.50 \\
    CNN-Transformer & 5.17 \\
    WF-2M + Linear & 5.15 \\
    WF-8M + Linear & 5.27 \\
    WF-2M + Transformer & \textbf{4.83} \\
    WF-8M + Transformer & 4.84 \\
    \bottomrule
\end{tabular}
    \label{tab:pretrained-soybeans-yield}
\end{table}
\subsection{Influenza Forecasting}
\begin{table}[htb]
    \centering
    \caption{Comparison of Validation MAE for ILI (\%) forecasting. The target variable had a mean of 2.43\% and a standard deviation of 1.73.}
    \begin{tabular}{p{0.33\textwidth}p{0.1\textwidth}p{0.1\textwidth}p{0.12\textwidth}}
    \toprule
    \textbf{Model} & \textbf{+1 Week MAE} & \textbf{+5 Weeks MAE} & \textbf{+10 Weeks MAE} \\\midrule
    Linear Regression & 0.392 & 0.437 & 0.456\\
    ARIMA          & 0.219 & 0.409 & 0.427\\
    Transformer (without weather)             & 0.193 & 0.328 & 0.371\\
    Transformer             & 0.189 & 0.294 & 0.320\\
    WF-2M + Transformer  &\textbf{0.186} & \textbf{0.277} & \textbf{0.297}\\
    WF-8M + Transformer  & 0.188 & 0.291 & 0.329\\ %
    \bottomrule
\end{tabular}
    \label{tab:pretrained-flu-pred}
\end{table}
In \autoref{tab:pretrained-flu-pred}, we observe that short-term influenza-like illness (ILI) behavior is predominantly influenced by recent influenza cases, as shown by the performance of the ARIMA models. However, a transformer encoder without weather can detect autoregressive features better, leading to better performance. Adding weather data improves the performance of the transformer for medium and long-term predictions. Additionally, \wf{}+Transformer-2M model shows considerable improvement over the base transformer model, further emphasizing the importance of good feature extraction. We also notice that the 8M pretrain model shows little to no improvement over the baseline transformer, suggesting that the model overfitted the small dataset. 

\section{Conclusion and Limitations}
We pretrained two \wf{} models and showed that fine-tuning them improved performance in crop yield prediction and influenza forecasting. However, several limitations merit discussion. Firstly, the pretraining dataset could be expanded to include a larger area and ground-level measurements. The model hyperparameters can be tuned with an exhaustive grid search. And lastly, the model needs to be retrained every year on new weather data to stay relevant. We aim to address these limitations in future work.


\bibliography{references}
\bibliographystyle{plainnat}

\newpage
\section*{Supplementary Material}
\section{Detailed Description of Datasets}
\label{sec:datasets}
\subsection{Pretraining Dataset}
\begin{figure}[htb]
    \centering
    \includegraphics[width=0.8\textwidth]{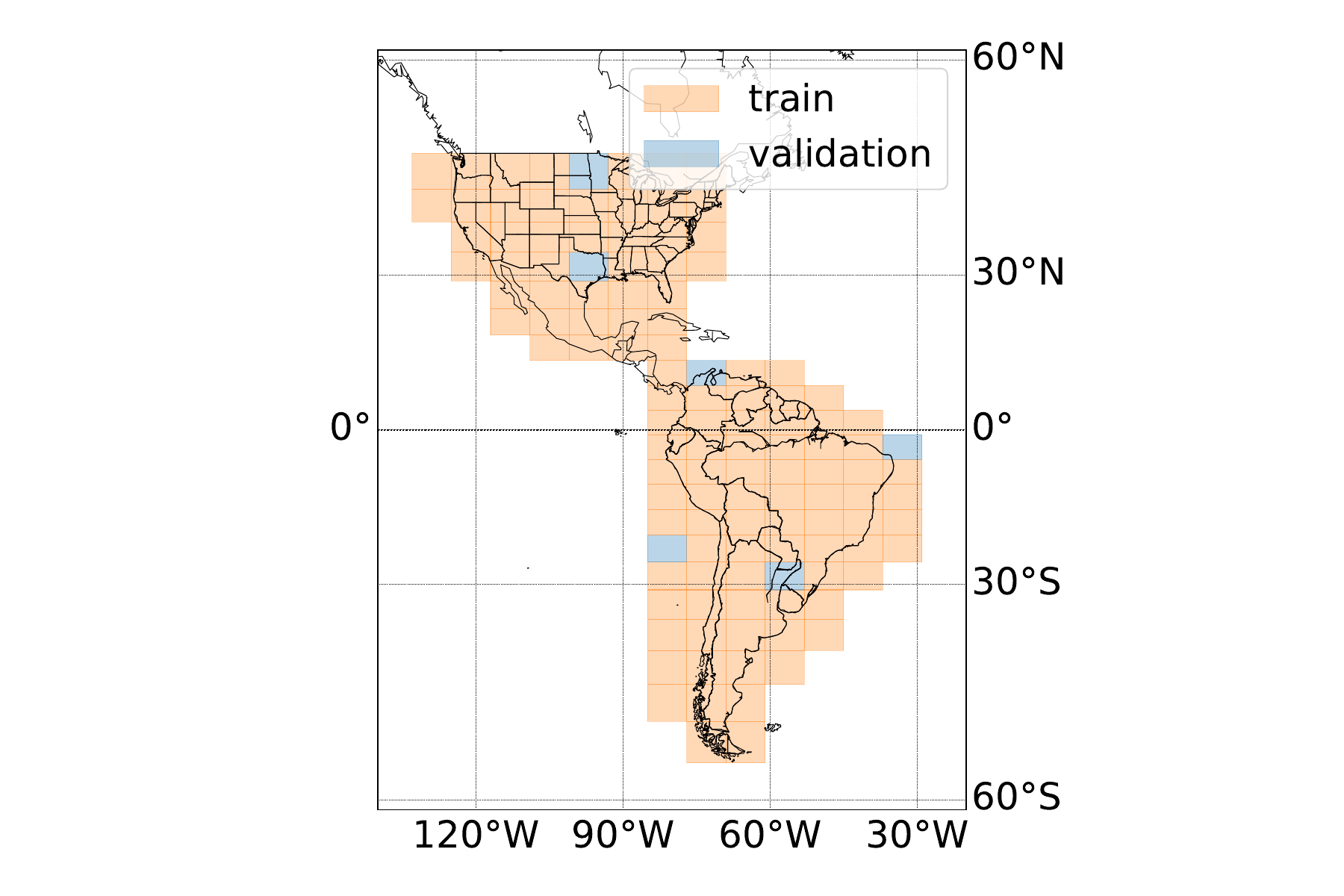}
    \caption{Map of the pretraining data. 5\% of the grid rectangles were selected at random for validation and the rest were used for pretraining the model.}
    \label{fig:pretraining-loc-map}
\end{figure}
The dataset contained a small percentage of missing values, which were subsequently imputed using data from preceding years. After data collection, we applied the Tetens equation \citep{Tetens1930} and the FAO Penman-Monteith formula \citep{Ndulue2021}, to calculate additional variables of interest such as saturation vapor pressure, vapor pressure deficit, and reference evapotranspiration. A complete list of all the weather measurements is given in \autoref{tab:weather_variables}. 

Additionally, we also computed weekly and monthly averages of the weather variables and added them to the datasets. Consequently, the \textsc{WeatherFormer} is trained on daily, weekly, and monthly granularities. The pretraining dataset in total had approximately $9.1$ billion floating point numbers. We used $95\%$ of the data for pretraining and $5\%$ data for validation as shown in \autoref{fig:pretraining-loc-map}.
\subsection{Soybean Yield Dataset}
\subsubsection*{Data Collection}
For the crop yield prediction problem, we used a soybean yield dataset introduced by \citet{khaki2019cnn}. This dataset contains county-level soybean yield for the nine states in the corn belt region of the United States from 1980 to 2018. Alongside soybean yield data, it also contains six weather variables (precipitation, solar radiation, snow water equivalent, maximum and minimum temperatures, and vapor pressure), soil profile across six depths for 10 characteristics (composition, bulk density, and organic matter content, etc) and management practices. Notably, the original paper reported results for 13 states and for corn and soybeans, but the publicly available dataset comprises data for only 9 states and only soybeans. Weather and practice data are provided weekly, whereas soil data lack a temporal component. 

\subsubsection*{Train-Validation Split}
The experiment was run five times and during each run, the data from randomly chosen seven states were used for training and the remaining two states were used for validation. The best Root Mean Square Error (RMSE) for the validation dataset from each of the five runs is averaged and reported.

It is important to note that the weather measurement source and granularity in this dataset are different from the pretraining data for \textsc{WeatherFormer}, and despite that the model showed considerable improvement over the baseline, suggesting a good understanding of the weather from pretraining.

\subsection{Influenza Dataset}
Another downstream predictive task we chose for the \textsc{WeatherFormer} was Influenza forecasting. The spread and severity of the influenza epidemic are strongly influenced by the mean temperature. Other weather variables such as humidity, precipitation, and wind speed also affect the process to varying degrees. Previously, several researchers, such as \citet{amendolara2023lstm} and \citet{yang2023deep} have used weather as a data source for prediction. We decided to use only the mean temperature from \citet{NASAPower} since our preliminary investigations on small models suggested that additional weather measurements introduce noise in the dataset.

\subsubsection*{Data Collection}
We obtained our influenza dataset from \citet{farrow2015delphi}, which includes weekly totals of Influenza-like Illness (ILI) cases and the percentage of ILI cases for New York City. The dataset spans 11 influenza seasons, from 2010-2011 to 2021-2022, and comprises weekly counts of patients, influenza cases, and the percentage of patients with ILI. We focused on predicting the percentage of ILIs among all patients. The dataset started on the week 40 of 2010. Due to a distribution shift attributed to COVID-19, data post-2020 were excluded. 

For weather, we used our satellite-based weather data averaged over New York City. \citet{amendolara2023lstm} reported that temperature is a strong predictor of influenza cases followed by precipitation and wind speed. Average cooling and heating days are also important as these metrics relate to the time people spend indoors, a factor influencing transmission. We, however, discovered that only average temperature was sufficient for good prediction. Adding more measurements did not seem to improve the performance. This could be because our data source was from the satellite, which might be too imprecise for the task.

\subsubsection*{Train-Validation Split}
We divided the remaining data into training and validation sets in four sequential phases: data from 2010 to 2015 was used for training and data from 2016 was used for validation. This pattern was repeated with training extended by one year each time, and validation occurring in the subsequent year. This resulted in four training and validation datasets. The best model's performance on the validation sets was averaged. Following the methodology of \citet{amendolara2023lstm} and others in the field, we reported Mean Absolute Error (MAE) as our evaluation metric.

We evaluated each model using a rolling forecasting approach.  Models received weather data and historical ILI cases for a fixed number of past weeks and were then tasked with predicting ILI cases for the subsequent 10 weeks without access to future weather information. Following each prediction cycle, the input window was shifted forward by one week, generating a new 10-week forecast that extended the previous one. This process was repeated which yielded 52 prediction tasks per year, each spanning 10 weeks.

We standardized all input data before training. We found that the historical window of input data critically influenced model performance. Consequently, we optimized this hyperparameter by exploring values within the range of ${105, 110, 115, \ldots, 135}$ weeks. We also found that for the ARIMA model, the performance is given by $p=54$, $d=1$ and $q=1$.

\section{Detailed Results for Finetuning Tasks}
\begin{table}[htb]
    \centering
    \caption{Details of soybean yield prediction across across five validation sets. The dataset mean and standard deviation were 38.5 Bu/Acre and 11.03 Bu/Acre, respectively. The best performance in each dataset is highlighted.}
    \begin{tabular}{p{0.3\textwidth}ccccc}
    \toprule
    & \multicolumn{5}{c}{\textbf{Validation RMSE}}\\
    \textbf{Model} & \textbf{Dataset 1} & \textbf{Dataset 2} & \textbf{Dataset 3} & \textbf{Dataset 4} & \textbf{Dataset 5} \\
    \midrule
    
    Linear Regression   & 5.52  & 7.28  & 8.44  & 7.05  & 6.21\\
    CNN-RNN             & 4.74  & 5.14  & 6.94  & 5.37  & 5.31\\
    CNN-Transformer     & 4.58  & 5.38  & 6.12  & 5.08  & 4.68\\
    WF-2M + Linear      & 4.46  & 5.45  & 6.58  & 4.76  & 4.50\\
    WF-8M + Linear      & 4.49  & 5.38  & 6.91  & 4.85  & 4.72\\
    WF-2M + Transformer & \textbf{4.13}  & 5.16  & \textbf{5.88}  & 4.79  & \textbf{4.17} \\
    WF-8M + Transformer & 4.17  & \textbf{5.01}  & 5.89  & \textbf{4.75}  & 4.37 \\
    \bottomrule
\end{tabular}
    \label{tab:pretrained-soybeans-yield-details}
\end{table}

We present the detailed results for the soybean yield prediction task in \autoref{tab:pretrained-soybeans-yield-details}. Every time 7 states were chosen at random for training and the remaining two were chosen for validation. The validation state pairs were (South Dakota, Iowa), (Nebraska, Minnesota), (North Dakota, Kansas), (North Dakota, Minnesota), and (North Dakota, Missouri). 

Next, we present the detailed results for the influenza 
prediction in \autoref{tab:pretrained-flu-details}. We observe that ILI in 2018 had a sharp peak (\autoref{fig:nyc-flu-cases}), which made the performance of all the models worse but in the other years, the model performances were more consistent. 

\begin{longtable}{p{0.33\textwidth}p{0.1\textwidth}p{0.1\textwidth}p{0.12\textwidth}}
    \caption{Details of influenza forecasting evaluation across four validation sets. The dataset mean and standard deviation were 2.43\% and 1.73, respectively.}\\
    \toprule
    \label{tab:pretrained-flu-details}
    \textbf{Model} & \textbf{+1 Week MAE} & \textbf{+5 Weeks MAE} & \textbf{+10 Weeks MAE} \\\midrule
    \multicolumn{4}{c}{\textbf{2016}}\\\midrule
    Linear Regression & 0.318 & 0.290 & 0.350\\
    ARIMA          & 0.175 & 0.302 & 0.328\\
    Transformer (without weather)             & 0.133 & 0.246 & 0.291\\
    Transformer             & \textbf{0.124} & 0.243 & 0.245\\
    WF-2M + Transformer  & 0.127 & \textbf{0.224} & \textbf{0.233}\\
    WF-8M + Transformer  & 0.130 & 0.231 & 0.275\\\midrule 
    
    \multicolumn{4}{c}{\textbf{2017}}\\\midrule
    Linear Regression & 0.341 & 0.356 & 0.350\\
    ARIMA          & 0.179 & 0.297 & 0.298\\
    Transformer (without weather)             & \textbf{0.172} & 0.272 & 0.301\\
    Transformer             & 0.181 & 0.213 & 0.222\\
    WF-2M + Transformer  & 0.178 & \textbf{0.209} & \textbf{0.195}\\
    WF-8M + Transformer  & 0.181 & 0.224 & 0.234\\\midrule
    
    \multicolumn{4}{c}{\textbf{2018}}\\\midrule
    Linear Regression & 0.655 & 0.705 & 0.770\\
    ARIMA          & 0.333 & 0.717 & 0.683\\
    Transformer (without weather)             & \textbf{0.266} & 0.573 & 0.643\\
    Transformer             & 0.273 & 0.530 & 0.558\\
    WF-2M + Transformer  & 0.267 & \textbf{0.481} & \textbf{0.502}\\
    WF-8M + Transformer  & 0.273 & 0.502 & 0.527\\\midrule 
    
    \multicolumn{4}{c}{\textbf{2019}}\\\midrule
    Linear Regression & 0.256 & 0.398 & 0.354\\
    ARIMA          & 0.188 & 0.321 & 0.398\\
    Transformer (without weather)             & 0.199 & 0.223 & 0.249\\
    Transformer             & 0.180 & \textbf{0.188} & \textbf{0.255}\\
    WF-2M + Transformer  & 0.173 & 0.192 & 0.256\\
    WF-8M + Transformer  & \textbf{0.166} & 0.208 & 0.282\\
    \bottomrule
    \end{longtable}

\clearpage
\section{Ablation Studies}
\subsection{Pretraining}
We believe it is important to test whether the performance improvement in \autoref{tab:pretrained-soybeans-yield} and \autoref{tab:pretrained-flu-pred} are due to the utilization of transformer architecture in weather feature extraction or due to pertaining. We hypothesize that the latter is true. To verify this hypothesis, we trained the same models from scratch, without loading the pretraining weights. \autoref{tab:no-pretraining-soybeans-yield} and \autoref{tab:no-pretraining-flu-pred} show that all models perform worse without pretraining.
\begin{table}[htb]
    \centering
    \caption{Comparison of Validation RMSE for county-level soybeans yield forecasting without pertaining. The large models overfit the training data and perform terribly on the validation dataset.}
    \begin{tabular}{p{0.45\textwidth}l}
        \toprule
        \textbf{Model} & \textbf{Validation RMSE} \\\midrule
        Linear Regression & 6.90 \\
        \wf{}-2M + Transformer & 6.56 \\
        \wf{}-8M + Transformer & 6.61 \\
        \bottomrule
    \end{tabular}
    \label{tab:no-pretraining-soybeans-yield}
\end{table}

\begin{table}[htb]
    \centering
    \caption{Comparison of Validation MAE for ILI (\%) forecasting. The larger models show worse performance due to overfitting.}
    \begin{tabular}{p{0.33\textwidth}p{0.1\textwidth}p{0.1\textwidth}p{0.12\textwidth}}
    \toprule
    \textbf{Model} & \textbf{+1 Week MAE} & \textbf{+5 Weeks MAE} & \textbf{+10 Weeks MAE} \\\midrule
    Linear Regression    & 0.392 & 0.437 & 0.456\\
    WF-2M + Transformer  & 0.188 & 0.295 & 0.318\\
    WF-8M + Transformer  & 0.197 & 0.375 & 0.436\\
    \bottomrule
\end{tabular}
    \label{tab:no-pretraining-flu-pred}
\end{table}

\subsection{Training with Masked Language Modeling}
We trained two additional \wf{} models of size 2M and 8M respectively with Masked Language Modeling. Following BERT, 15\% of the tokens were masked with zero. We observe that the 8M parameter model struggles to fit the pretraining data with MLM. Previously, it was able to fit the data with Masked Feature Prediction and feature swapping, which resulted in the pretrained \wf{}-8M model used in previous experiements. 

We tested different learning rates between 0.0005 and 0.001 and warmup periods but did not show improvement, suggesting a limitation of the MLM approach for pretraining on weather data.

\begin{figure}[tbh]
    \centering
    \begin{subfigure}{0.45\textwidth}
    \includegraphics[width=\textwidth]{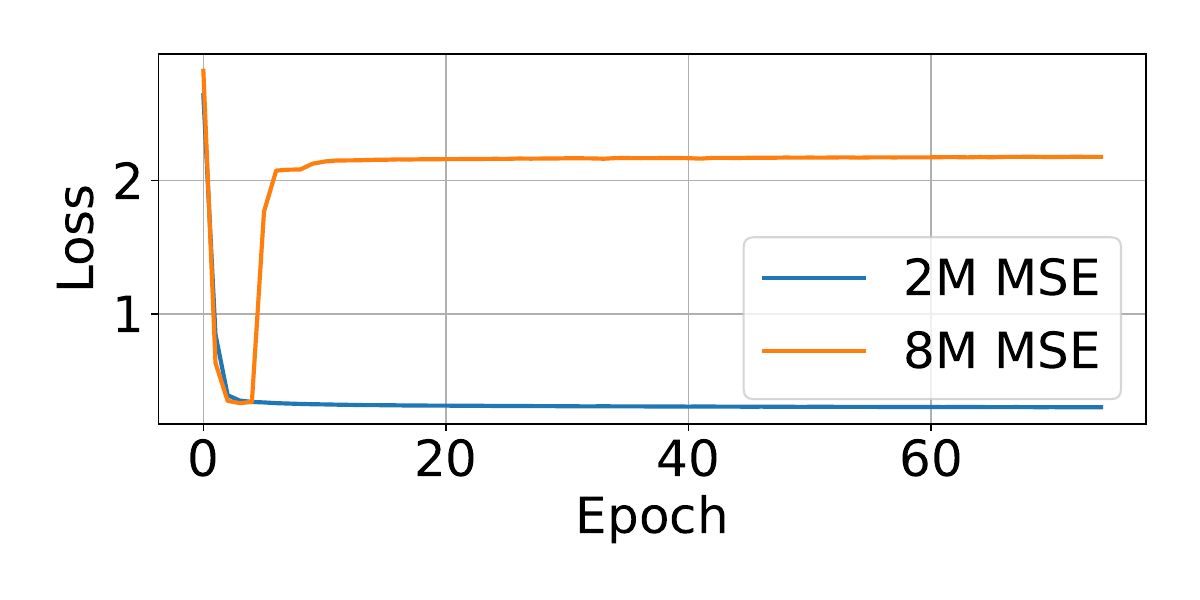}
    \caption{Training.}
    \label{fig:bert-pretraining-train-loss}
    \end{subfigure}
    \begin{subfigure}{0.45\textwidth}
    \includegraphics[width=\textwidth]{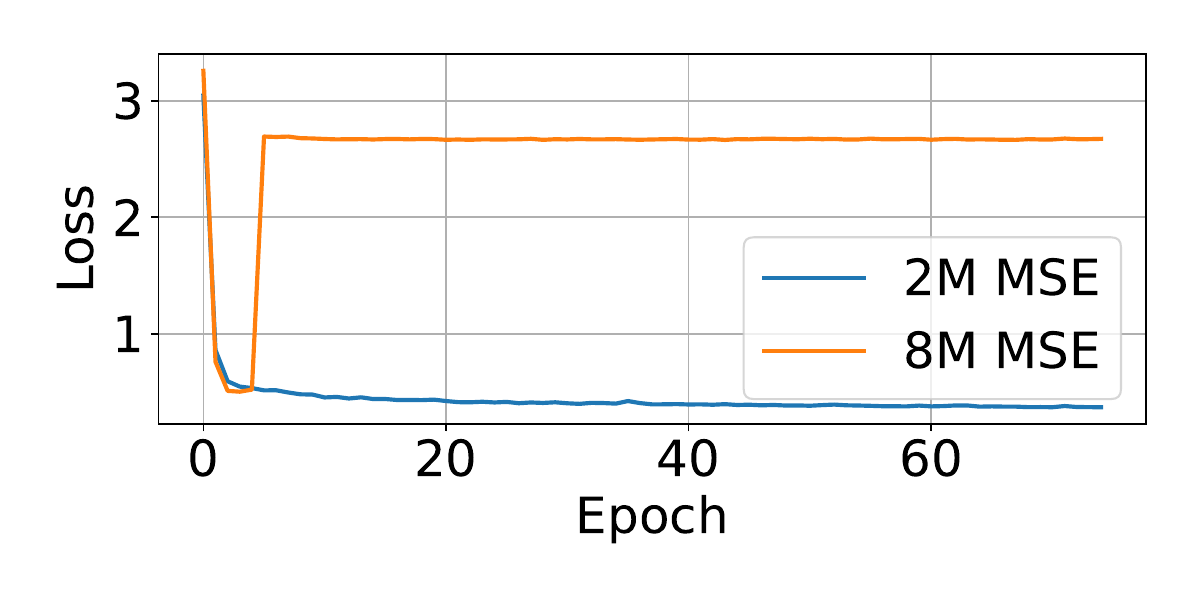}
    \caption{Validation}
    \label{fig:bert-pretraining-val-loss}
    \end{subfigure}
     \caption{Comparison between the 2M and 8M model losses during pretraining with MLM. The 8M model failed to converge.}
    \label{fig:bert-pretraining-loss}
\end{figure}

\section{Meteorological Equations}
Weather measurements adhere to the principles of physics. When certain weather measurements are unavailable, they can be estimated using established meteorological equations. By leveraging these equations, we estimated additional atmospheric properties using collected data.

\subsection{Tetens Equation}
The Tetens equation, as described by \citet{Tetens1930}, offers a straightforward method to estimate the saturation vapor pressure ($e_a$) over liquid water or ice, contingent on the ambient temperature. For temperatures above freezing ($x > 0^\circ C$, representing liquid water), the equation is:
\[
e_a = 0.6108 \cdot \exp\left(\frac{17.27 \cdot x}{x + 237.3}\right)
\]

Conversely, for temperatures at or below freezing ($x \leq 0^\circ C$, representing ice), the equation modifies to:
\[
e_a = 0.6108 \cdot \exp\left(\frac{21.87 \cdot x}{x + 265.5}\right)
\]

In these equations, $e_a$ represents the saturation vapor pressure in kPa, and $x$ denotes the temperature in degrees Celsius ($^\circ C$).

\subsection{FAO Penman-Monteith Equation}
Accurate estimation of evapotranspiration (ET) is crucial for water resource management, agricultural planning, and understanding the hydrological cycle. The FAO Penman-Monteith equation \citep{Ndulue2021} is widely accepted as the standard method for calculating reference evapotranspiration ($ET0$) from climatic data. The equation is expressed as:
\[
ET0 = \frac{0.408 \Delta (R_n - G) + \gamma \frac{900}{T + 273} u_2 (e_s - e_a)}{\Delta + \gamma (1 + 0.34 u_2)}
\]

where:
\begin{itemize}
    \item $ET0$ is the reference evapotranspiration (mm day$^{-1}$),
    \item $\Delta$ is the slope of vapor pressure curve (kPa $^\circ$C$^{-1}$),
    \item $R_n$ is the net radiation at the crop surface (MJ m$^{-2}$ day$^{-1}$),
    \item $G$ is the soil heat flux density (MJ m$^{-2}$ day$^{-1}$),
    \item $T$ is the mean daily air temperature at 2 m height ($^\circ$C),
    \item $u_2$ is the wind speed at 2 m height (m s$^{-1}$),
    \item $e_s$ is the saturation vapor pressure (kPa),
    \item $e_a$ is the actual vapor pressure (kPa),
    \item $\gamma$ is the psychrometric constant (kPa $^\circ$C$^{-1}$).
\end{itemize}

This equation assumes a standard grass reference crop with an assumed height of 0.12 m, a fixed surface resistance of 70 sm$^{-1}$, and an albedo of 0.23.

\section{List of Weather Measurements in the Pretraining Data}
Here is a list of 31 daily weather measurements in the pretraining data. The first 28 measurements were downloaded from the NASA Power Project \citep{NASAPower} from 1984 to 2022. The last three were predicted using the above equations. 

\begin{longtable}{p{0.4\textwidth}ll}
\caption{Descriptions of the 31 weather measurements and their units.}\\\toprule
\label{tab:weather_variables}
\textbf{Measurement Name} & \textbf{Symbol}           & \textbf{Unit}               \\\midrule
Temperature at 2 Meters                                       & T2M                       & $^\circ$C                   \\ 
Temperature at 2 Meters Maximum                               & T2M\_MAX                   & $^\circ$C                   \\ 
Temperature at 2 Meters Minimum                               & T2M\_MIN                   & $^\circ$C                   \\ 
Wind Direction at 2 Meters                                    & WD2M                      & Degrees                     \\ 
Wind Speed at 2 Meters                                        & WS2M                      & m/s                         \\ 
Surface Pressure                                              & PS                        & kPa                         \\ 
Specific Humidity at 2 Meters                                 & QV2M                      & g/Kg                        \\ 
Precipitation Corrected                                       & PRECTOTCORR               & mm/day                      \\ 
All Sky Surface Shortwave Downward Irradiance                 & ALLSKY\_SFC\_SW\_DWN       & MJ/m$^2$/day                \\ 
Evapotranspiration Energy Flux                                & EVPTRNS                   & MJ/m$^2$/day                \\ 
Profile Soil Moisture (0 to 1)                                & GWETPROF                  & 0 to 1                      \\ 
Snow Depth                                                    & SNODP                     & cm                          \\ 
Dew/Frost Point at 2 Meters                                   & T2MDEW                    & $^\circ$C                   \\ 
Cloud Amount                                                  & CLOUD\_AMT                & 0 to 1                      \\ 
Evaporation Land                                              & EVLAND                    & kg/m$^2$/s $\times 10^6$    \\ 
Wet Bulb Temperature at 2 Meters                              & T2MWET                    & $^\circ$C                   \\ 
Land Snowcover Fraction                                       & FRSNO                     & 0 to 1                      \\ 
All Sky Surface Longwave Downward Irradiance                  & ALLSKY\_SFC\_LW\_DWN       & MJ/m$^2$/day                \\ 
All Sky Surface PAR Total                                     & ALLSKY\_SFC\_PAR\_TOT      & MJ/m$^2$/day                \\ 
All Sky Surface Albedo                                        & ALLSKY\_SRF\_ALB           & 0 to 1                      \\ 
Precipitable Water                                            & PW                        & cm                          \\ 
Surface Roughness                                             & Z0M                       & m                           \\ 
Surface Air Density                                           & RHOA                      & kg/m$^3$                    \\ 
Relative Humidity at 2 Meters                                 & RH2M                      & 0 to 1                      \\ 
Cooling Degree Days Above 18.3 C                              & CDD18\_3                   & days                        \\ 
Heating Degree Days Below 18.3 C                              & HDD18\_3                   & days                        \\ 
Total Column Ozone                                            & TO3                       & Dobson units                \\ 
Aerosol Optical Depth 55                                      & AOD\_55                    & 0 to 1                      \\ 
Reference Evapotranspiration                                  & ET0                       & mm/day                      \\ 
Vapor Pressure                                                & VAP                       & kPa                          \\ 
Vapor Pressure Deficit                                        & VAD                       & kPa                          \\\bottomrule
\end{longtable}

\end{document}